\documentclass[aps,prx,twocolumn,superscriptaddress,a4paper,english,longbibliography]{revtex4-2}

\usepackage[utf8]{inputenc}
\usepackage[T1]{fontenc}
\usepackage{graphicx}
\graphicspath{{./figures/}}
\usepackage{subfigure} 
\usepackage{amsmath, amssymb, amsfonts}
\usepackage{hyperref}
\usepackage{multirow}
\usepackage{array}
\usepackage{booktabs}
\usepackage{xcolor}

\newcommand{\R}[1]{\textcolor{black}{#1}}
\newcommand{\B}[1]{\textcolor{black}{#1}}
\newcommand{\RR}[1]{\textcolor{black}{#1}}

\begin{document}

\title{Compressing \RR{Neural Networks Using Tensor Networks with Exponentially Fewer Variational Parameters}}

\author{Yong Qing}
\thanks{These authors contributed equally to this work.}
\affiliation{Department of Physics, Capital Normal University, Beijing 100048, China.}

\author{Ke Li}
\thanks{These authors contributed equally to this work.}
\affiliation{Department of Physics, Capital Normal University, Beijing 100048, China.}

\author{Peng-Fei Zhou}
\affiliation{Department of Physics, Capital Normal University, Beijing 100048, China.}

\author{Shi-Ju Ran}
\email{sjran@cnu.edu.cn}
\affiliation{Department of Physics, Capital Normal University, Beijing 100048, China.}
	
\begin{abstract}
		Neural network (NN) designed for challenging machine learning tasks is in general a highly nonlinear mapping that contains massive variational parameters. High complexity of NN, if unbounded or unconstrained, might unpredictably cause severe issues including \R{overfitting}, loss of generalization power, and unbearable cost of hardware. In this work, we propose a general compression scheme that significantly reduces the variational parameters of NN's, despite of their specific types (linear, convolutional, \textit{etc}), by encoding them to deep \R{automatically differentiable} tensor network (ADTN) that contains exponentially-fewer free parameters. Superior compression performance of our scheme is demonstrated on several widely-recognized NN's (FC-2, LeNet-5, AlextNet, ZFNet and VGG-16) and datasets (MNIST, CIFAR-10 and CIFAR-100). For instance, we compress two linear layers in VGG-16 with approximately $10^{7}$ parameters to two ADTN's with just 424 parameters, improving the testing accuracy on CIFAR-10 from $90.17\%$ to $91.74\%$. We argue that the deep structure of ADTN is an essential reason for the remarkable compression performance of ADTN, compared to  existing compression schemes that are mainly based on tensor decompositions/factorization and shallow tensor networks. Our work suggests deep TN as an exceptionally efficient mathematical structure for representing the variational parameters of NN's, which exhibits superior compressibility over the commonly-used matrices and multi-way arrays.
\end{abstract}
	
\maketitle

\section{Introduction}

Neural network (NN)~\cite{DLNN15R} has achieved remarkable success across a wide range of fields including computer vision, natural language processing, and most recently scientific investigations in, e.g., applied mathematics (e.g., ~\cite{AGVOTGM19,AMGHAI21}) and  physics (e.g., ~\cite{SQMBPNN17,ERQMSDNN17,MLPOM17,NNRG18,SSMVAN19PRL,MLPS19,MTR21MLHamilt}). To enhance performance in handling complex real-world tasks, such as human interaction (e.g.,~\cite{NEURIPS2022_b1efde53}) and robot control (e.g.,\cite{singh_efficient_2023,singh_deep_2023,singh_efficient_2024}), the amount of variational parameters in NN rapidly increased from millions to trillions (e.g., GPT-3\R{,} with 175 billion parameters~\cite{LMAFSL20gtp3}). Such a paradigm with the utilization of super large-scale NN's brought us several severe challenges. Though the representation ability of NN should be enhanced by increasing complexity, R{overfitting} might occur and the generalization ability of NN might unpredictably be harmed. The increasing complexity also brings unbearable hardware costs in academic investigations and practical applications.

The variational parameters in NN are usually stored as matrices or higher-order tensors (also called \R{multilinear} arrays). Previous results show that generalization ability can be improved by suppressing or ``refining'' the degrees of freedom in such tensors by, e.g., network pruning~\cite{OPD1989MC,RVNP2019NP,CNNP21}, knowledge distillation~\cite{DKNN2015,KDAS2021}, weight sharing~\cite{SNNSWS1992,REFWSNAS21}, tensor R{decompositions and factorizations}~\cite{EUTNDCNN19,DNNCNN22}, \textit{etc}.

Among the inspiring progresses made in recent years, a simple and efficient iterative approach known as MUSCO\cite{gusak_automated_2019} alternates the low-rank factorization with smart rank selection and fine-tuning. Mixed tensor decomposition (MTD)~\cite{liang_automatic_2021} attempts to combine Tucker decomposition and canonical polyadic decomposition (CPD) to for better performance. Low-rank adaptation (LoRA) \cite{hu2021loralowrankadaptationlarge} freezes all pre-trained parameters and inserts a trainable pair of matrices (acting as a low-rank decomposition of a full matrix) additively into each layer of the Transformer architecture.

Meanwhile, tensor network (TN) demonstrates its significant prospects in model compression. Matrix product state (MPS)~\cite{PVWC07MPSRev, GCEMPO2017}, which is also known as tensor-train (TT) form~\cite{TTD11}, and matrix product operator (MPO)~\cite{MPOR10}, were applied to compress the tensors in NN's, and achieved remarkable compression ratios~\cite{TNN15,CDCNN15MPSCNN,CDNNBMPO20,MCMMPO20,WCTRN18}. Nonlinear TT~\cite{wang_nonlinear_2021} was introduced by containing nonlinear activation functions embedded in the sequenced contractions and convolutions on the top of the normal TT decomposition. These imply TN as a superior mathematical structure over simple \R{multilinear} arrays for representing the variational parameters of NN's, which yet remains an open issue. However, these works mainly concern to use the shallow TN's. The utilization of deep TN's, which were revealed to be essentially different and more powerful from the research in quantum physics and machine learning, has not been investigated yet for model compression.

\begin{figure}[tpb]
	\centering
	\includegraphics[angle=0,width=1\linewidth]{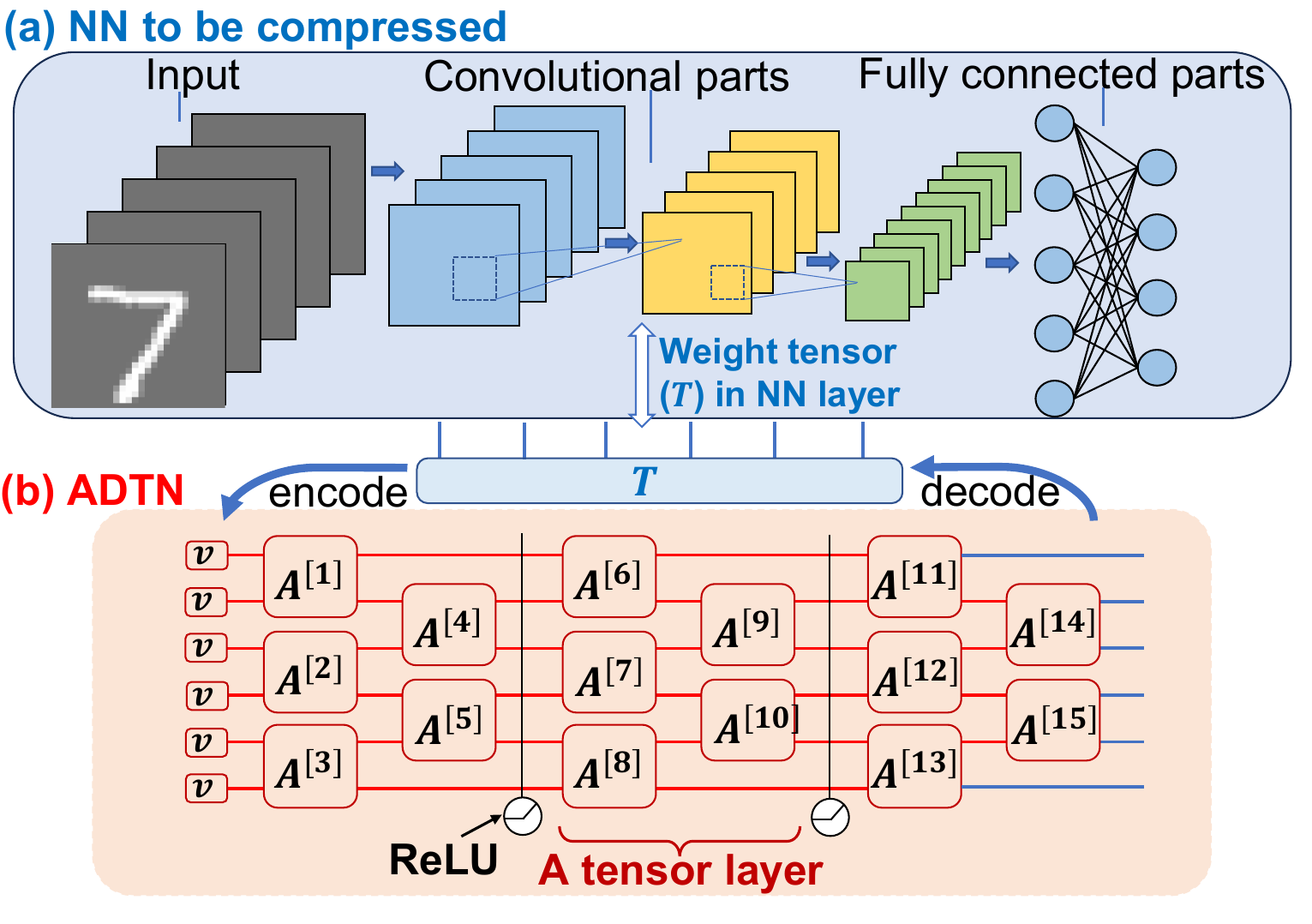}
	\caption{(Color online) The workflow of ADTN for compressing NN. (a) The illustration of a convolutional NN as an example, whose variational parameters ($\boldsymbol{T}$) are encoded in a ADTN shown in (b). The contraction of the ADTN results in $\boldsymbol{T}$, in other words, where the ADTN contains much less parameters than $\boldsymbol{T}$.}
	\label{fig-1}
\end{figure}

In this work, we propose to encode the variational parameters of NN layer(s) into the contraction of TN~\cite{RTPC+17TNrev} that contains exponentially fewer parameters. See the illustration of the main procedures in Fig.~\ref{fig-1}. For instance, the number of parameters of the TN encoding $2^Q$ parameters of NN scale just linearly as $O(MQ)$, with $M\sim O(1)$ the number of TN layers. Since the contraction process is differentiable, automatic differentiation technique~\cite{LLWX19ADTN,ZHR21ADQC} is utilized to optimize the tensors of the TN to reach the optimal accuracy after compression. Thus, we dub our scheme as automatically differentiable tensor network (ADTN).

ADTN exhibits fundamental differences from previous compression methods. Unlike traditional approaches that rely on explicit \R{decomposition or factorization} procedures or the insertion of matrices, ADTN directly encodes NN parameters into a deep TN and updates them through automatic differentiation techniques. This distinguishes ADTN from the TT-based methods and LoRA. By utilizing deep TN's, ADTN leverages a distinct representational power that surpasses the capabilities of shallow TN's used in the existing tensor \R{decompositions, factorizations}, and MPS/MPO methods. We will give more discussions later in Sec. 4.

We demonstrate the compression performance of ADTN on various well-known NN's (FC-2, LeNet-5~\cite{GBLADR1998LCY}, AlexNet, ZFnet, and VGG-16~\cite{VDCN14VGG16}) and datasets (MNIST~\cite{MNIST_web},  CIFAR-10 and CIFAR-100~\cite{CIFAR10}). Note that the details of NN's can be found in Appendix A. For instance, in the experiment of compressing two fully connected layers of VGG-16, we compressed about $10^7$ parameters to two ADTN's with just 424 parameters, where the testing accuracy of CIFAR-10 improved from $90.17\%$ to $91.74\%$. Our work implies a general way to reduce the excessive complexity of NN's by encoding the variational parameter into TN's, which serve as a more compact and efficient mathematical form.

\section{Methods}\label{sec:QML}

\subsection{\R{Encoding variational parameters of NN's to ADTN's}}

An NN can be formally written as a mapping
\begin{align}
	\boldsymbol{y} = f(\boldsymbol{x}; \boldsymbol{T}, \boldsymbol{T}', \ldots),
	\label{eq-f}
\end{align}
which maps the input sample $\boldsymbol{x}$ to the output vector $\boldsymbol{y}$. We aim to compress the variational parameters $(\boldsymbol{T}, \boldsymbol{T}', \ldots)$ or a part of them by ADTN. Note we do not assume the types of layers (linear, convolutional, \textit{etc}.) in the NN as it makes no difference to our ADTN method. 

Denoting the parameters to be compressed as tensor $\boldsymbol{T}$, our idea is to represent $\boldsymbol{T}$ as the contraction of ADTN with multiple layers. The structure of ADTN is flexible. Here, we choose the commonly used ``brick-wall'' structure; see an example of its diagrammatic representation in Fig.~\ref{fig-1} (b). In simple words about the diagrammatic representation of TN, each block represents a tensor, and the bonds connected to a block represent the indexes of the corresponding tensor. Shared bonds represent the dumb indexes that should be summed up (contracted). The vertical black lines represent the activation function (e.g., ReLU). More details about TN contractions and the diagrammatic representation can be found in Appendix A or Ref.~\cite{RTPC+17TNrev}. 

For simplicity, we assume to compress $2^Q$ variational parameters, i.e., $\#(\boldsymbol{T}) = 2^Q$, into one ADTN. Obviously, the contraction of ADTN should also give us a tensor (denoted as $\boldsymbol{\mathcal{T}}$) containing $2^Q$ parameters. In our example, the ADTN results in a $Q$-th order tensor, where the dimension of each index is taken as $d=2$. These indexes are represented by the unshared bonds in the ADTN. See the blue bonds on the right boundary of the diagrammatic representation in Fig.~\ref{fig-1} (b).

The exemplified ADTN is formed by several $(d \times d \times d \times d)$ tensors $\{\boldsymbol{A}^{[k]}\}$ ($k=1, \ldots, K$ with $K$ as their total number), which form the variational parameters of ADTN. On the left boundary of ADTN, we put several two-dimensional constant vectors $\boldsymbol{v}$ for the convenience in formulating and analyzing. We may take $\boldsymbol{v} = [1, 0]$ for instance.

For the brick-wall structure, we define each $(Q-1)$ tensors in two neighboring columns as a TN layer [depicted in Fig.~\ref{fig-1} (b)]. We put an activation function between each two neighboring TN layers. Contracting the tensors in one tensor layer essentially gives us a ($2^Q \times 2^Q$) matrix denoted as $\boldsymbol{\mathcal{L}}^{(m)}$ for the $m$-th tensor layer. The mapping represented by the whole ADTN can be written as
\begin{align}
	\boldsymbol{\mathcal{T}} =\boldsymbol{\mathcal{L}}^{(M)} \sigma( \ldots \boldsymbol{\mathcal{L}}^{(3)} \sigma(\boldsymbol{\mathcal{L}}^{(2)} \sigma(\boldsymbol{\mathcal{L}}^{(1)} \prod_{\otimes q=1}^Q \boldsymbol{v}))),
	\label{eq-Uv}
\end{align}
with $\otimes$ the tensor product and $\sigma$ the activation function. In other words, $\boldsymbol{\mathcal{T}}$ is obtained by implementing the mappings $\{\sigma (\boldsymbol{\mathcal{L}}^{(m)}(\ast))\}$ (for $m=1, 2, \cdots, M$ with $M$ the number of TN layers) on $\prod_{\otimes q=1}^Q \boldsymbol{v}$. 

Be aware that one does not have to compute the matrices $\{\boldsymbol{\mathcal{L}}^{(m)}\}$, which are introduced here only for the convenience of understanding and expressing. Instead, an efficient strategy for contracting an ADTN involves sequentially contracting the 4th-order tensors $\{\boldsymbol{A}^{[k]}\}$ from left to right with $\prod_{\otimes q=1}^Q \boldsymbol{v}$. Note that we assume the tensors $\{\boldsymbol{A}^{[k]}\}$ to be $(2 \times 2 \times 2 \times 2)$ in this work, though their dimensions can be chosen flexibly. The computational complexity scales approximately as $O(2^{Q+2}K)$. Note that contracting one $4$-th order tensor is equivalent to multiplying a $(2^{Q-2} \times 4)$ matrix with a $(4 \times 4)$ matrix, with a complexity of $O(2^{Q+2})$.

By encoding $2^{Q}$ parameters, the total number of variational parameters in the brick-wall ADTN depicted in Fig.~\ref{fig-1} (b) scales as
\begin{align}
	\# (\text{ADTN}) \equiv \sum_k \#(\boldsymbol{A}^{[k]}) \sim O(MQ),
	\label{eq-complex1}
\end{align}
with $M \sim O(1)$ according to our results provided below. In plain words, we lower the exponential complexity $O(2^Q)$ to linear complexity $O(MQ)$ with respect to $Q$.

\subsection{\R{Optimization of ADTN}}

There are two main stages to obtain the tensors \R{$\{\boldsymbol{A}^{[k]}\}$ in an ADTN}. The first is pre-training by minimizing the distance between $\boldsymbol{T}$ in NN and  $\boldsymbol{\mathcal{T}}$ obtained from ADTN. We choose the loss function as Euclidean distance $L^{\text{E}} = |\boldsymbol{\mathcal{T}} - \boldsymbol{T}|$\R{, as the parameter values of the neural network lack specific meaning. For more on the impact of different pretraining strategies, see Appendex \ref{app-C}.}

After $L^{\text{E}}$ converges, we move to the second stage to minimize the loss function for implementing the machine learning task. One may choose the same loss function as the one used for training the NN, such as the cross entropy from the outputs of NN and the ground-truth classifications of the training samples. The feed-forward process of getting the output of the NN is the same as that for training the NN, except that the parameters ($\boldsymbol{T}$) of the target NN layer are replaced by the tensor ($\boldsymbol{\mathcal{T}}$) obtained by contracting the ADTN. 

Generally speaking, the first stage is to enhance the stability by finding a proper initialization of ADTN, and the second stage further improves the performance with an end-to-end optimization. The mapping given by the NN in the second stage can be written as
\begin{align}
	\boldsymbol{y} = f(\boldsymbol{x}; \{\boldsymbol{A}^{[k]}\}, \boldsymbol{T}^{\text{res}}),
	\label{eq-f1}
\end{align}
with $\boldsymbol{T}^{\text{res}}$ denoting the uncompressed parameters.

The number of compressed NN layers, the number of ADTN's, and the number of TN layers within the ADTN's are all flexible. The mapping of the compressed NN can be expressed as $\boldsymbol{y} = f(\boldsymbol{x}; \{\boldsymbol{A}^{[k]}\}, \{\boldsymbol{B}^{[k]}\}, \allowbreak \cdots, \boldsymbol{T}^{\text{res}})$, with $\{\boldsymbol{A}^{[k]}\}, \{\boldsymbol{B}^{[k]}\}, \cdots$ denoting the tensors of multiple ADTN's. This flexibility facilitates handling cases where the number of parameters cannot be expressed as $d^Q$. For instance, $(3\times 2^{10})$ parameters in a NN layer can be compressed into a single ADTN with $Q=11$ unshared indexes, where 10 are two-dimensional and one is three-dimensional. Alternatively, these parameters can also be divided into two parts with $2^{11}$ and $2^{10}$ parameters, respectively, which are then compressed to two ADTN's with $Q=11$ and $10$, respectively. All unshared indexes in these two ADTN's can be two-dimensional.

\R{Let us provide a concrete example to explain the choice of $Q$ and the number of ADTNs. We may consider the compression of a $(600 \times 600)$ weight matrix (denoted as $\boldsymbol{W}$) in a linear layer. One can consider compressing a $(512 \times 512)$ part of $\boldsymbol{W}$, which contains $2^{18} = 262144$ parameters (with $Q = 18$). This leaves $360000 - 262144 = 97856$ parameters uncompressed, where one can optionally introduce another ADTN of $Q = 16$ to compress $2^{16} = 65536$ parameters. Following \B{this approach}, \B{an appropriate} number of ADTNs can be introduced to \B{ensure} that most (\B{or a desired portion) of a given parameter tensor is compressed into ADTNs}. Another way to compress the \B{entire} $\boldsymbol{W}$ with a single ADTN is to \B{choose} $Q = 19$, with $2^{19} = 524288 > 360000$. One \B{could} add zeros to $\boldsymbol{W}$ to increase the number of parameters \B{to $2^{19}$}. But notably, the number of parameters in the $Q=19$ ADTN will still be much lower than that of $\boldsymbol{W}$.}

\RR{ADTN distinguishes itself from conventional compression techniques such as network distillation, model quantization~\cite{nagel2021white}, and low-rank decompositions. Network distillation~\cite{OPD1989MC,RVNP2019NP,CNNP21} transfers knowledge from a larger model to a smaller one, while model quantization reduces numerical precision to minimize memory and computational costs. Low-rank decompositions ~\cite{EUTNDCNN19,DNNCNN22} approximate high-dimensional data by representing it with lower-rank approximations, thereby reducing parameter complexity. In contrast, ADTN directly encodes the parameter tensors into deep TN's. The ADTN method is independent of network distillation and model quantization, though these techniques can be integrated into ADTN’s pre-training and fine-tuning stages to further enhance compression. ADTN can be considered a lower-rank structure with a deep architecture, distinguishing it from the relatively shallow structures typically produced by conventional low-rank decompositions.}

\section{Results and Discussions}

\subsection{Compression Performance}
For all the ADTN's used in this work, we take the brick-wall structure, where all tensors are $(2 \times 2 \times 2 \times 2)$. The activation functions are taken as ReLU. The value of $Q$ is taken according to the size of the compressed tensor. We only vary the number of ADTN's ($N$) and the number of TN layers ($M$) in each ADTN in the compression of each NN layer. The specific values of $N$ and $M$ are given in the corresponding texts describing the results.

\begin{table*}[t]
\centering
\setlength{\tabcolsep}{3pt}
\renewcommand{\arraystretch}{1.2} 

\resizebox{0.8\linewidth}{!}{    \begin{tabular}{l|c|c|cc|c|c|ccc}
        \hline
                   & \multirow{2}{*}{Dataset}   & \multicolumn{1}{c|}{\multirow{2}{*}{\begin{tabular}[c]{@{}c@{}}NN\\ Model\end{tabular}}} & \multicolumn{2}{c|}{\#Parameters}                                                                                                                                                 & \multicolumn{1}{c|}{\multirow{2}{*}{\begin{tabular}[c]{@{}c@{}}Compression \\ ratio\\ $\rho=\mathcal{P}/P$\end{tabular}}} & \multicolumn{1}{c|}{\multirow{2}{*}{\begin{tabular}[c]{@{}c@{}}\#ADTN\\ $N$\end{tabular}}} & \multicolumn{3}{c}{Testing accuracy}                                                                                                 \\ \cline{4-5} \cline{8-10} 
                   &                            & \multicolumn{1}{c|}{}                                                                    & \multicolumn{1}{c|}{\begin{tabular}[c]{@{}c@{}}Compressed\\ NN layer(s) $P$\end{tabular}} & \multicolumn{1}{c|}{\begin{tabular}[c]{@{}c@{}}ADTN(s) \\ $\mathcal{P}$\end{tabular}} & \multicolumn{1}{c|}{}                                                                                                  & \multicolumn{1}{c|}{}                                                                      & \multicolumn{1}{c|}{\begin{tabular}[c]{@{}c@{}}Baseline\\ $\eta_{NN}$\end{tabular}} & \multicolumn{1}{c|}{$\eta$} & $\eta/\eta_{NN}$ \\ \hline

        \multirow{9}{*}{\rotatebox{90}{Compress \textbf{linear} layers(s)}} 
        & \multirow{2}{*}{MNIST} & FC-2 & \( 131072 \) & \( 160 \) & \( 1.2 \times 10^{-3} \) & \( 1 \) & \( 97.94\% \) & \( 97.81\% \) & \( 99.69\% \) \\ 
        & & LeNet-5 & \( 8192 \) & \( 120 \) & \( 1.5 \times 10^{-2} \) & \( 1 \) & \( 99.15\% \) & \( 99.48\% \) & \( 100.36\% \) \\ 
        \cline{2-10}
        & \multirow{4}{*}{CIFAR-10} & LeNet-5 & \( 163840 \) & \( 300 \) & \( 1.8 \times 10^{-3} \) & \( 2 \) & \( 74.39\% \) & \( 75.27\% \) & \( 101.18\% \) \\ 
        & & AlexNet & \( 5242880 \) & \( 388 \) & \( 7.4 \times 10^{-5} \) & \( 2 \) & \( 81.13\% \) & \( 81.31\% \) & \( 100.22\% \) \\ 
        & & ZFNet & \( 6291456 \) & \( 404 \) & \( 6.4 \times 10^{-5} \) & \( 2 \) & \( 80.59\% \) & \( 81.03\% \) & \( 100.66\% \) \\ 
        & & VGG-16 & \( 18874368 \) & \( 424 \) & \( 2.2 \times 10^{-5} \) & \( 2 \) & \( 90.17\% \) & \( 91.74\% \) & \( 101.74\% \) \\ 
        \cline{2-10}
        & \multirow{3}{*}{CIFAR-100} & LeNet-5 & \( 163840 \) & \( 300 \) & \( 1.8 \times 10^{-3} \) & \( 2 \) & \( 41.30\% \) & \( 42.60\% \) & \( 103.15\% \) \\ 
        & & AlexNet & \( 5242880 \) & \( 388 \) & \( 7.4 \times 10^{-5} \) & \( 2 \) & \( 49.80\% \) & \( 49.19\% \) & \( 98.78\% \) \\ 
        & & ZFNet & \( 6291456 \) & \( 404 \) & \( 6.4 \times 10^{-5} \) & \( 2 \) & \( 50.19\% \) & \( 50.36\% \) & \( 100.34\% \) \\ 
        \hline
        \multirow{8}{*}{\rotatebox{90}{Compress \textbf{conv} layers(s)}} 
        & MNIST & LeNet-5 & \( 32768 \) & \( 140 \) & \( 4.2 \times 10^{-3} \) & \( 1 \) & \( 99.15\% \) & \( 99.52\% \) & \( 100.37\% \) \\ 
        \cline{2-10}
        & \multirow{4}{*}{CIFAR-10} & LeNet-5 & \( 8192 \) & \( 120 \) & \( 1.5 \times 10^{-2} \) & \( 1 \) & \( 74.39\% \) & \( 75.16\% \) & \( 101.04\% \) \\ 
        & & AlexNet & \( 1572864 \) & \( 364 \) & \( 2.9 \times 10^{-4} \) & \( 2 \) & \( 81.13\% \) & \( 81.09\% \) & \( 99.95\% \) \\ 
        & & ZFNet & \( 393216 \) & \( 324 \) & \( 8.2 \times 10^{-4} \) & \( 2 \) & \( 80.59\% \) & \( 81.63\% \) & \( 101.29\% \) \\ 
        & & VGG-16 & \( 11796480 \) & \( 1820 \) & \( 1.5 \times 10^{-4} \) & \( 10 \) & \( 90.17\% \) & \( 91.09\% \) & \( 101.02\% \) \\ 
        \cline{2-10}
        & \multirow{3}{*}{CIFAR-100} & LeNet-5 & \( 8192 \) & \( 120 \) & \( 1.5 \times 10^{-2} \) & \( 1 \) & \( 41.30\% \) & \( 42.54\% \) & \( 103.00\% \) \\ 
        & & AlexNet & \( 1572864 \) & \( 364 \) & \( 2.9 \times 10^{-4} \) & \( 2 \) & \( 49.80\% \) & \( 50.15\% \) & \( 100.70\% \) \\ 
        & & ZFNet & \( 393216 \) & \( 324 \) & \( 8.2 \times 10^{-4} \) & \( 2 \) & \( 50.19\% \) & \( 50.88\% \) & \( 101.37\% \) \\ 
        \hline
    \end{tabular}}
\caption{ADTN's performance on compressing the linear layers (green area) and convolutional layers (orange area) of FC-2, LeNet-5, Alex Net, ZF Net, and VGG-16 for MNIST, CIFAR-10, and CIFAR-100 datasets. We show the number of parameters in the compressed NN layer(s) $P$, that of ADTN(s) $\mathcal{P}$, compression ratio $\rho = \mathcal{P}/P$, the number of ADTN(s) $N$, and testing accuracies before compression $\eta_{\text{NN}}$ and after compression $\eta$. The hyper-parameters of these NN's can be found in Appendix B.}
\label{tab:table1}
\end{table*}

In Table~\ref{tab:table1}, the green area shows the performance of ADTN with $M=1$ for compressing linear layers in various NN's. High compression ratios $\rho = \mathcal{P}/P$ (with $P$ and $\mathcal{P}$ the number of compressed parameters in the NN layer(s) and that of the ADTN(s), respectively) are achieved. 

Our first example is FC-2 formed by two linear layers, whose sizes are $(784 \times 256)$ and $(256 \times 10)$, respectively. We slice out $2^{17}$ parameters from the first linear layer and compress them in an ADTN with just 160 parameters. The testing accuracy on the MNIST dataset, which consisting of 60000 hand-written images of size $(28 \times 28)$~\cite{MNIST_web}, remains almost unchanged after compression. 

In more complicated convolutional NN's (LeNet-5, AlexNet, and ZFNet), we chose to compress all linear layers except for the output layer. This is because the number of parameters in the output layer is much smaller compared to other layers. For different datasets (MNIST, CIFAR-10, and CIFAR-100), we just accordingly alter the dimensions for the input and output of each NN, and keep other hyper-parameters unchanged. Note CIFAR-10 and CIFAR-100 contain totally 80 million everyday-life RGB images of size $(32 \times 32)$~\cite{CIFAR10}. 

The ADTN's with just hundreds of parameters manage to compress the linear layers whose numbers of parameters vary from $O(10^{3})$ to $O(10^{7})$, while the testing accuracy is not harmed by compression. Remarkably in compressing VGG-16, we use two ADTN's with in total 424 parameters to compress two linear layers with approximately $10^{7}$ parameters (whose dimensions are $(512 \times 4096)$ and $(4096 \times 4096)$, respectively), and meanwhile are able to increase the testing accuracy from $90.17\%$ (baseline) to $91.74\%$.

The orange area in Table~\ref{tab:table1} shows the excellent performance of ADTN to compress convolutional layers. For example, we compress the two largest convolutional layers in AlexNet and ZFNet (with about $O(10^{6})$ and $O(10^{5})$ parameters) that account for more than $80\%$ of the total parameters in the corresponding NN. These parameters are compressed to ADTN's consisting of just 324 parameters. For VGG-16, we choose to compress the five largest layers from the $13$ convolutional layers. Each convolutional layer used two ADTN's to compress. The testing accuracy is improved for about 1 percentage.

We shall emphasize that many settings of our ADTN compression scheme are flexible, such as the choices of compressed NN layers, the number of ADTN's, and the hyper-parameters of each ADTN. The demonstrated results in this paper are obtained after slightly adjusting these quantities. Higher compression performance can generally be reached by adjusting them more carefully.

\begin{figure*}[htb]
    \begin{minipage}{0.2\textwidth}
        \centering
        \includegraphics[width=0.9\textwidth]{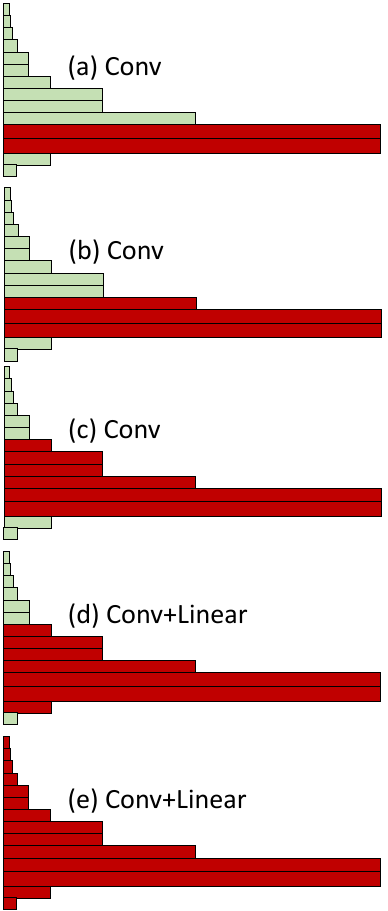}
        \label{fig:example}
    \end{minipage}
    \hfill
    \begin{minipage}{0.7\textwidth}

\resizebox{0.66\linewidth}{!}{\begin{tabular}{c|cccccc}
\hline
    \begin{tabular}[c]{@{}c@{}}Compressed\\ layers(red)\end{tabular}           & M & $\tilde{\eta}$ & $\eta$    & $\eta /\eta_{NN}$ & $\rho$               & $\rho_{tot}$             \\ \hline
\multirow{3}{*}{(a) Conv}                                                   & 1 & $99.87\%$      & $84.27\%$ & $103.86\%$        & $5.0 \times 10^{-5}$ & \multirow{3}{*}{$0.435$} \\
                                                                           & 3 & $100\%$        & $85.52\%$ & $105.40\%$        & $2.1 \times 10^{-4}$ &                          \\
                                                                           & 5 & $100\%$        & $85.71\%$ & $105.63\%$        & $3.7 \times 10^{-4}$ &                          \\ \hline
\multirow{3}{*}{(b) Conv}                                                   & 1 & $99.99\%$      & $84.36\%$ & $103.97\%$        & $6.0 \times 10^{-5}$ & \multirow{3}{*}{$0.279$} \\
                                                                           & 3 & $100\%$        & $85.19\%$ & $105.00\%$        & $2.5 \times 10^{-4}$ &                          \\
                                                                           & 5 & $100\%$        & $85.68\%$ & $105.60\%$        & $4.4 \times 10^{-4}$ &                          \\ \hline
\multirow{3}{*}{(c) Conv}                                                   & 1 & $98.83\%$      & $78.53\%$ & $99.78\%$         & $8.0 \times 10^{-5}$ & \multirow{3}{*}{$0.135$} \\
                                                                           & 3 & $99.93\%$      & $82.44\%$ & $101.60\%$        & $3.3 \times 10^{-4}$ &                          \\
                                                                           & 5 & $100\%$        & $82.79\%$ & $102.03\%$        & $5.9 \times 10^{-4}$ &                          \\ \hline
\multirow{4}{*}{\begin{tabular}[c]{@{}c@{}}(d) Conv\\ +Linear\end{tabular}} & 1 & $96.35\%$      & $65.93\%$ & $81.25\%$         & $1.0 \times 10^{-4}$ & \multirow{4}{*}{$0.060$} \\
                                                                           & 3 & $99.94\%$      & $77.73\%$ & $95.78\%$         & $4.2 \times 10^{-4}$ &                          \\
                                                                           & 5 & $100\%$        & $79.15\%$ & $97.55\%$         & $7.4 \times 10^{-4}$ &                          \\
                                                                           & 7 & $100\%$        & $80.61\%$ & $99.35\%$         & $1.1 \times 10^{-3}$ &                          \\ \hline
\multirow{4}{*}{\begin{tabular}[c]{@{}c@{}}(e) Conv\\ +Linear\end{tabular}} & 3 & $75.78\%$      & $72.13\%$ & $88.90\%$         & $1.0 \times 10^{-4}$ & \multirow{4}{*}{$0.008$} \\
                                                                           & 5 & $78.13\%$      & $74.24\%$ & $91.50\%$         & $4.2 \times 10^{-4}$ &                          \\
                                                                           & 7 & $81.54\%$      & $77.61\%$ & $95.65\%$         & $7.4 \times 10^{-4}$ &                          \\
                                                                           & 9 & $81.95\%$      & $78.59\%$ & $96.86\%$         & $1.1 \times 10^{-3}$ &                          \\ \hline
\end{tabular}}
        \label{tab:example}
    \end{minipage}
\caption{(Color online) The performance of ADTN scheme with different numbers of tensor layers ($M$) to compress both the linear and convolutional layers of a reduced version of VGG-16 (with the baseline testing accuracy $\eta_{\text{NN}}=81.14\%$). In the figure at the left side, the red bars indicate the compressed layers, and their lengths illustrate the numbers of parameters, while the green bars represent the uncompressed layers. In the table at the right side, the three and four columns show the training and testing accuracies ($\tilde{\eta}$ and $\eta$, respectively). The last column gives the total compression ratio $\rho_{\text{tot}}$ [Eq.~(\ref{eq-rhot})].}
\label{VGG-16}
\end{figure*}

One can consider to simultaneously compress multiple types of NN layers. In Table~\ref{VGG-16}, we show the performance of ADTN for simultaneously compressing linear and convolutional layers of VGG-16. We use a reduced version of VGG-16 (with smaller sizes of the NN layers and without data augmentation) for demonstration, with its baseline testing accuracy $\eta_{\text{NN}}=81.14\%$. \RR{In the figure at the left side of Table ~\ref{VGG-16}, the red bars indicate the compressed
layers, and their lengths illustrate the numbers of parameters, while the green bars represent the
uncompressed layers.} Here, each NN layer is compressed to one ADTN. Remarkable compression ratios ($\rho$) are achieved.

In the last column, we show the total compression ratio
\begin{align}
	\rho_{\text{tot}} = \frac{\#(\text{ADTN}) + \#(\boldsymbol{T}^{\text{res}})}{\#(\text{NN})} \simeq \frac{\#(\boldsymbol{T}^{\text{res}})}{\#(\text{NN})}. 
	\label{eq-rhot}
\end{align}
The approximation holds since in general we have $ \#(\text{ADTN}) \ll \#(\boldsymbol{T}^{\text{res}}) \ll \#(\text{NN})$. Therefore, $\rho_{\text{tot}}$ can be significantly improved by simply compressing more NN layers to ADTN's. 

Our results suggest that for $\rho \lesssim 0.3$, shallow ADTN (say with $M=1$ tensor layer) is sufficient for compressing the chosen parts of the NN. The training accuracy $\tilde{\eta}$ is almost $100\%$, meaning the representational ability of the shallow ADTN is sufficient. As $\rho$ decreases by compressing more NN layers, $\tilde{\eta}$ also decreases. Deeper ADTN with better representational ability is required. As shown in the last row, the testing accuracy is improved from $72.13\%$ to $78.95\%$ by increasing $M$ from $3$ to $9$ when compressing 12 layers in VGG-16. The total compression ratio is lowered to $\rho_{\text{tot}} \simeq 0.008$.

Table~\ref{tabel-compare} compare the performances of ADTN with various schemes, including Tucker decomposition, CPD, MUSCO, and MTD. We take the data reported in Refs.~[\cite{gusak_automated_2019}] and [\cite{liang_automatic_2021}] for comparison. Three typical results of ADTN are given in the lower half of Table~\ref{tabel-compare}. The mark ($N$-$M$) of ADTN indicates using $N$ ADTN's, each containing $M$ TN layers, for compression. $\mathcal{M}_{\text{NN}}$ is the number of compressed NN layers, with totally 16 layers in VGG-16. ADTN demonstrates remark compression performances. With ($6$-$3$), our ADTN scheme compressing 8 NN layers achieves the highest testing accuracy $\eta = 91.90\%$, with the total compressions ratio $\rho_{tot} = 1/17.65$. This clearly surpasses Tucker decomposition and CPD. Additionally, with ($6$-$3$) for compressing 11 NN layers, much better total compression ratio $\rho_{\text{tot}} = 1/73.47$ is achieved, which far surpasses the total compression ratios obtained by the existing methods.

\begin{table}[htbp]
	
	\centering
	\setlength{\tabcolsep}{6pt}
    \resizebox{\linewidth}{!}{
	\begin{tabular}{c|cccc}
		\hline
		VGG-16                                               & $\mathcal{M}_{\text{NN}} $ & $1/\rho_{\text{tot}}$       & $\eta$           & $\eta/\eta_{NN}$                  \\ \hline
		Baseline~\cite{liang_automatic_2021}                 & -          & -            & 90.63\%          & -                            \\
		Tucker~\cite{liang_automatic_2021}                   & -                 & 11.82 & 91.02\%          & 100.43\%                     \\
		CPD~\cite{liang_automatic_2021}                      & -                    & 11.76 & 87.02\%          & 96.02\%                      \\
		MUSCO~\cite{gusak_automated_2019}             & -            & 23.55 & 87.07\%          & 96.07\%                      \\
		MTD~\cite{liang_automatic_2021}                    & -             & 32.59& 89.80\%          & 99.08\%                      \\ \hline
		Baseline (our)                        & -               & -            & 90.54\%          & -                            \\
		ADTN (2-3)                          &  10                & 46.97 & 89.85\%          & \multicolumn{1}{l}{99.24\%}  \\
		ADTN (6-3)                         & 8                  & 17.56 & \textbf{91.90\%} & \multicolumn{1}{l}{\textbf{101.50\%}} \\
		ADTN (6-3)                         & 11                  & \textbf{73.47} & 88.55\%          & \multicolumn{1}{l}{97.80\%}  \\ \hline
	\end{tabular}
    }
    \caption{The inverse total compression ratios $1/\rho_{\text{tot}}$, after-compression testing accuracy $\eta$, and the ratio against the baseline accuracy with various existing compression schemes and ADTN. $\mathcal{M}$ denotes the number of compressed NN layers by ADTN. The mark ($N$-$M$) specifies the number of ADTN's $N$ used for compressing one NN layer and the number of TN layers $M$ in each ADTN. The compressed NN in this table is VGG-16 by considering the CIFAR-10 dataset.}
	\label{tabel-compare}
\end{table}

\begin{figure}[htbp]
	\includegraphics[angle=0,width=1\linewidth]{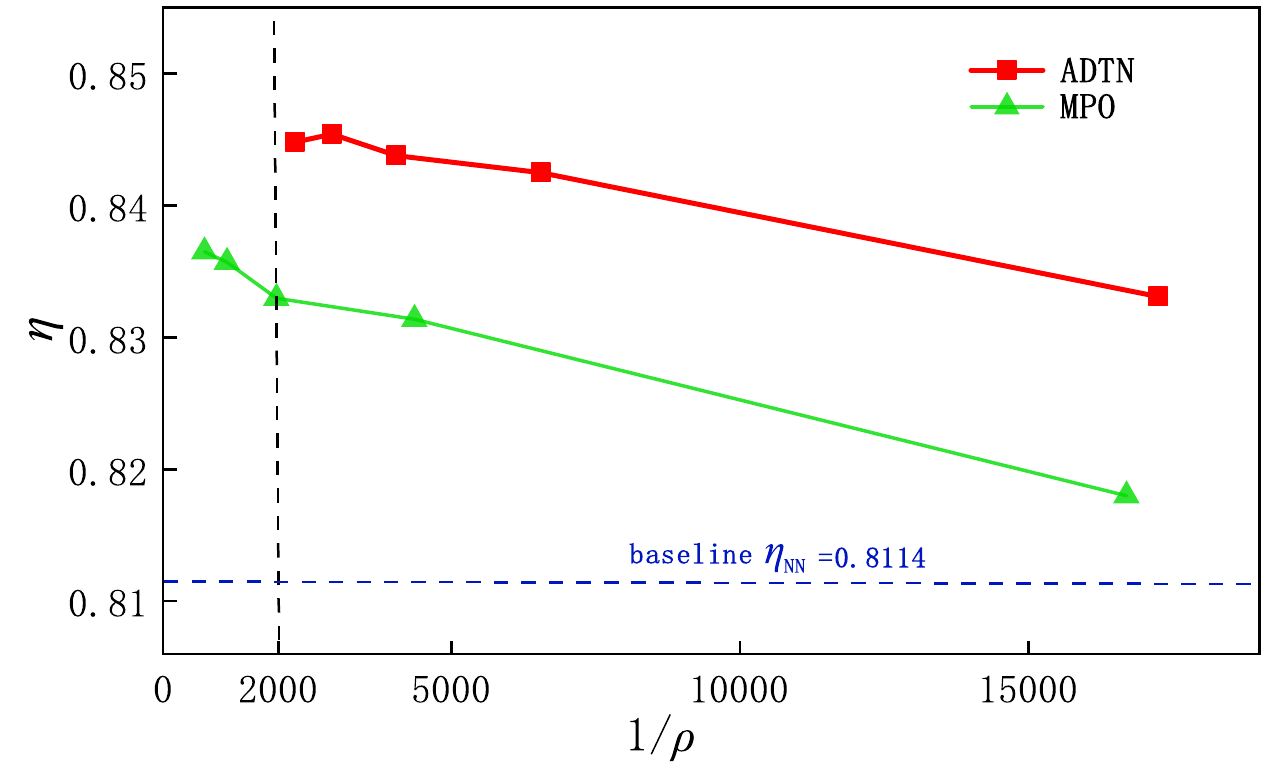}
	\caption{(Color online) The after-compression testing accuracy $\eta$ on CIFAR-10 versus the inverse compression ratio $1/\rho$, obtained by compressing a simplified VGG-16 using ADTN and MPO (shallow TN). The baseline testing accuracy is $\eta_{\text{NN}} = 0.8114$.}
	\label{fig-mpo}
\end{figure}

We also compare ADTN with the compression scheme based on MPO~\cite{CDNNBMPO20}, another typical shallow TN. MPO has a similar structure to MPS (TT form), and is used to represent quantum operators instead of states. Fig.~\ref{fig-mpo} shows the testing accuracy $\eta$ on CIFAR-10 after compressing the last convolutional layers in VGG-16 versus the inverse compression ratio $1/\rho$. The number of parameters of MPO is adjusted by varying its virtual bond dimensions. The number of TN layers in the ADTN ranges from $M=1$ to $5$. Both methods exceeds the baseline accuracy (horizontal dash line), while ADTN generally achieves higher inverse compression ratio. This is in accordance with previous results in the relevant fields, which will be discussed in Sec.~4.4.

\subsection{Discussions on compression with under/overfitting}

Below, we further discuss on several critical issues about the ADTN compression scheme. First, we demonstrate that by properly increasing the number of ADTN's ($N$), the NN can be effectively compressed whether it is under-fitting or overfitting. We take LeNet-5 as an example and alter the output dimension of the first linear layer $s$ (i.e., the input dimension of the second linear layer) to change the parameter complexity of the NN. Fig.~\ref{fig-VGG-16} (a) shows the testing-accuracy ratio $\eta/\eta_{\text{NN}}$ versus the inverse total compression ratio $\rho_{\text{tot}}^{-1}$  for the CIFAR-10 dataset with $s$ ranging from $32$ to $1536$.

The NN with a small $s$ is difficult to compress even by ADTN. In this case, the model is under-fitting, resulting in small $\rho_{\text{tot}}^{-1}$ and $\eta/\eta_{\text{NN}}$ (see the symbols with relatively light colors). By increasing $N$, $\eta/\eta_{\text{NN}}$ will fairly approach to $1$, mitigating the harm of accuracy. 

In contrast, for large $s$ where the NN is overfitting, the NN is shown to be compressible by ADTN, leading to large $\rho_{\text{tot}}^{-1}$ and $\eta/\eta_{\text{NN}}>1$, particularly when using multiple ADTN's for compression ($N>1$). See the symbols with dark colors. The testing accuracy is increased by compression with $\eta/\eta_{\text{NN}}>1$, implying that the generalization ability is improved. Generally, our results suggest that compressing by multiple ADTN's will avoid significant decrease of the testing accuracy in the under-fitting cases, and enhance generalization ability in the \R{overfitting} cases.

\begin{figure*}[htbp]
	\centering
 	\includegraphics[angle=0,width=0.9\linewidth]{L_V.pdf}
	\caption{(Color online) (a) The testing-accuracy ratio $\eta/\eta_{\text{NN}}$ versus the inverse total compression ratio $\rho_{\text{tot}}^{-1}$ for the CIFAR-10 dataset by LeNet-5. The dimensions of the first two linear layers are taken as $(512 \times s)$ and $(s \times 128)$, where $s$ is taken as $32, 64, \cdots, 1536$ (see the color bar). $N$ ADTN's are used for compression, where each contains $M=3$ TN layers. (b) The testing-accuracy ratio $\eta/\eta_{\text{NN}}$ versus the inverse of compression ratio $\rho^{-1}$ for the CIFAR-10 dataset by VGG-16. We selected $12$ largest layers (ten convolutional and two linear layers) in VGG-16 and compress them in two different orders: from the input to output (denoted as forward compression) and the other way around (denoted as backward compression).}
	\label{fig-VGG-16}
\end{figure*}

\subsection{Discussions on local-minima problem and compression order}

Considering compressing multiple layers in an NN, one suffers from severe local-minima problem. Simultaneously compressing several NN layers can be unstable\R{, as} optimization process may easily be trapped in different local minima when taking different initializations of the ADTN's. A stable way we have found is to compress layer by layer. 

Specifically, we start from the compression of one layer by encoding it as ADTN(s). After this is done, we compress another layer by optimizing the ADTN(s), and simultaneously optimize the ADTN(s) of the \R{formerly compressed} layer. For each time the optimization process converges, we \R{consider compressing} one more layer by optimizing the ADTN(s) of this layer and those of all optimized layers simultaneously. This requires \R{determining} a compression order.

We show that the compression order should be ``backward'', meaning compressing from the layers closer to the output of NN to those nearer the input. Fig.~\ref{fig-VGG-16} (b) shows the $\eta/\eta_{\text{NN}}$ versus $\rho^{-1}$ for the CIFAR-10 dataset by VGG-16. We compress 12 largest layers therein from the input to output (forward) and the other way around (backward). Significantly better performance is achieved with the backward compression. This is essentially due to the interdependence among the compressions of different layers. An accurate compression of a NN layer should require accurate input from the former layers in the information propagation process. We conjecture the above observation to be held when using other compression means.

\subsection{Discussions on Compression Faithfulness}
\begin{figure}[htbp]
	\centering
	\includegraphics[angle=0,width=1\linewidth]{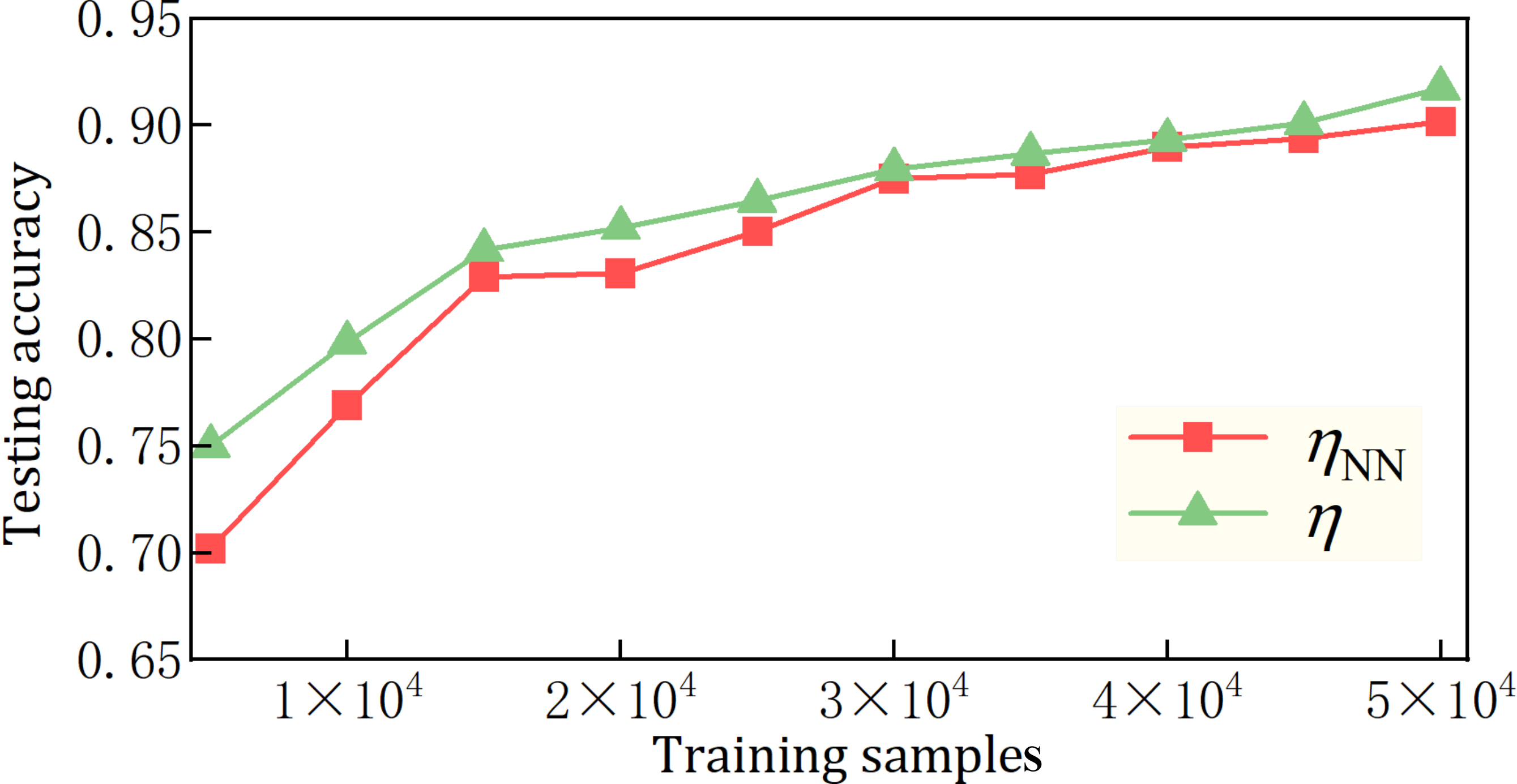}
	\caption{(Color online) The testing accuracy of NN $\eta_{\text{NN}}$ and that after compression $\eta$ of VGG-16 on the CIFAR-10 dataset with different numbers of training samples. The compression by our ADTN scheme faithfully restores the testing accuracy of the original NN with slight improvements.}
	\label{fig-Training sample}
\end{figure}

We expect a \R{well-performing} compression scheme to at least faithfully restore the generalization ability of the original NN, regardless of whether the NN itself is well-trained or not. This means the testing accuracy after compression will be (at least) close to the testing accuracy of the original NN (the baseline). Fig.~\ref{fig-Training sample} shows the testing accuracies before and after compression (denoted as $\eta_{\text{NN}}$ and $\eta$) for VGG-16 on the CIFAR-10 dataset. Different testing accuracies \R{before compression}  are obtained by varying the number of training samples. Our results show that the ADTN compression scheme faithfully restores the testing accuracies of the NN with slight improvements. We expect the ADTN scheme to be faithful for compressing other NN models as well. 

\subsection{Discussions on Tensor Network: ``deep'' versus ``shallow''}
\label{sec-TNDS}

Deep and shallow TN's possess essential differences in the fields of quantum physics, quantum computation, and quantum-inspired machine learning. Here, we specifically refer to the shallow TN's as those with just one layer. In quantum many-body physics, MPS (a shallow TN) can well approximate the gapped states of one-dimensional (1D) quantum systems~\cite{SWVC08MPSent}, where the entanglement entropy satisfies the so-called 1D area law. The projected entangled pair state (PEPS), which belongs to deep TN's, can approximate the states with much higher entanglement (obeying the two-dimensional area law)~\cite{F13arealawTRG, GE16arealawTNS}.

Note entanglement closely associates with the \R{nonsparsity} or more specifically\R{, the} number of \R{nonvanishing} singular \R{values and} thus closely associates with the compressibility. In other words, deep TN's can generally approximate quantum states whose coefficients form much denser tensors compared to those approximated by shallow TN's. Sparsity is closely related to compressibility. A tensor, whether it stores some data or variational parameters, possesses certain compressibility if it is sparse. This coincides with our results, where deep ADTN exhibits higher compression performance over the schemes based on MPS (TT form) and MPO. Similar observations have been made in the TN-based quantum-inspired machine learning~\cite{SR23QAI}\R{ where} PEPS outperforms MPS for classification and generation~\cite{CWZ21PEPSML}.

A more explicit example demonstrating the superior power of deep TN's over the shallow ones is the preparation of MPS by quantum circuit. An MPS can be encoded as the evolution of a product state (analogous to $\prod_{\otimes q=1}^Q \boldsymbol{v}$ in Eq.~(\ref{eq-Uv})) by implementing a quantum circuit formed by $O(1)$ layers of quantum gates (which is an unitary deep TN, analogous to our ADTN formed by several TN layers). Previous work showed that the number of parameters in a quantum circuit is generally much smaller than that in an MPS~\cite{ZHR21ADQC}, suggesting that an MPS can be further compressed by deep TN. Our work was inspired by this observation, and \R{experiments such as those in Fig.~\ref{fig-mpo} support} that ADTN's with moderate depths ($O(1)$ TN layers) possess superior compression abilities compared to MPS or MPO.

\section{Conclusion}
In summary, we proposed a general compression scheme based on automatically differentiable tensor network (ADTN) to significantly reduce the variational parameters of neural networks (NN's). The key idea is to encode the higher-order parameter tensors of NN as the contractions of deep ADTN(s) that contain exponentially fewer parameters. The performance of our scheme has been demonstrated with several well-known NN's on different datasets. Discussions on \R{overparametrization}, compression order, faithfulness, and depth of ADTN were given. Our work suggests that deep tensor network offers a more efficient and compact mathematical representation of the variational parameters in NN's compared to multi-way arrays and shallow tensor networks.

The high flexibility of the proposed ADTN scheme allows for further improvements in compression performance. Besides the tuning of \R{hyperparameters} (depth, bond dimensions, \textit{etc}.) of ADTN, architectures other than the brick-wall type can be considered to construct the ADTN. The activation functions can be replaced by others such as \R{leaky ReLU}. Instead of compressing one NN layer with multiple ADTN's, one might try to compress multiple NN layers into a single ADTN by concatenating their parameters into one tensor. This is a reflection of the generality of our scheme, and might be useful in special cases (e.g., when there exist redundant NN layers). Again, the types of NN layers make no difference to ADTN scheme.

Though ADTN performs impressively for compression, the decoding process (contraction) introduces additional computational cost to inference. This is because the contraction of ADTN and the inference of NN are currently treated as two separated procedures. One should first implement the contraction to get the parameters of NN, then followed by inference using the NN. Computational efficiency can be improved by combining these two procedures, which we leave for the future investigation.

\section*{Acknowledgments}

Y.Q. is thankful to Ze-feng Gao for stimulating discussions. 

\noindent\textbf{Funding}: This work was supported in part by Beijing Natural Science Foundation (Grant No. 1232025), the Ministry of Education Key Laboratory of Quantum Physics and Photonic Quantum Information (Grant No. ZYGX2024K020), and Academy for Multidisciplinary Studies, Capital Normal University.

\noindent\textbf{Author contributions}: S.J.R. conceived and initiated the presented ideas, and supervised the research. Y.Q. and K.L. contributed to most of the coding works and numerical simulations. All authors contributed to the algorithms, analyses, and the preparation of the manuscript. 

\noindent\textbf{Competing interests}: The authors declare that they have no competing interests.

\subsection*{Data availability}
The data are available on request from the corresponding authors.

%

\medskip
\clearpage
\newpage
\appendix
\section*{Appendix}
\section{Necessity of pre-training}
\label{app-C}

\begin{figure*}[htp]
	\centering
	\caption{\R{(Color online) Loss versus the training epochs in the second (fine-tuning) stage. The left panel shows the loss for ADTN(2-8) with three conditions: no pre-training, pre-training using L2 (Euclidean) distance, and pre-training using L1 distance. By ``(2-8)'', we mean to take $2$ ADTN's for compressing each NN layer, where each ADTN contains 8 tensor layers. The right panel presents the losses for ADTN(4-2).}}
	\includegraphics[angle=0,width=1\linewidth]{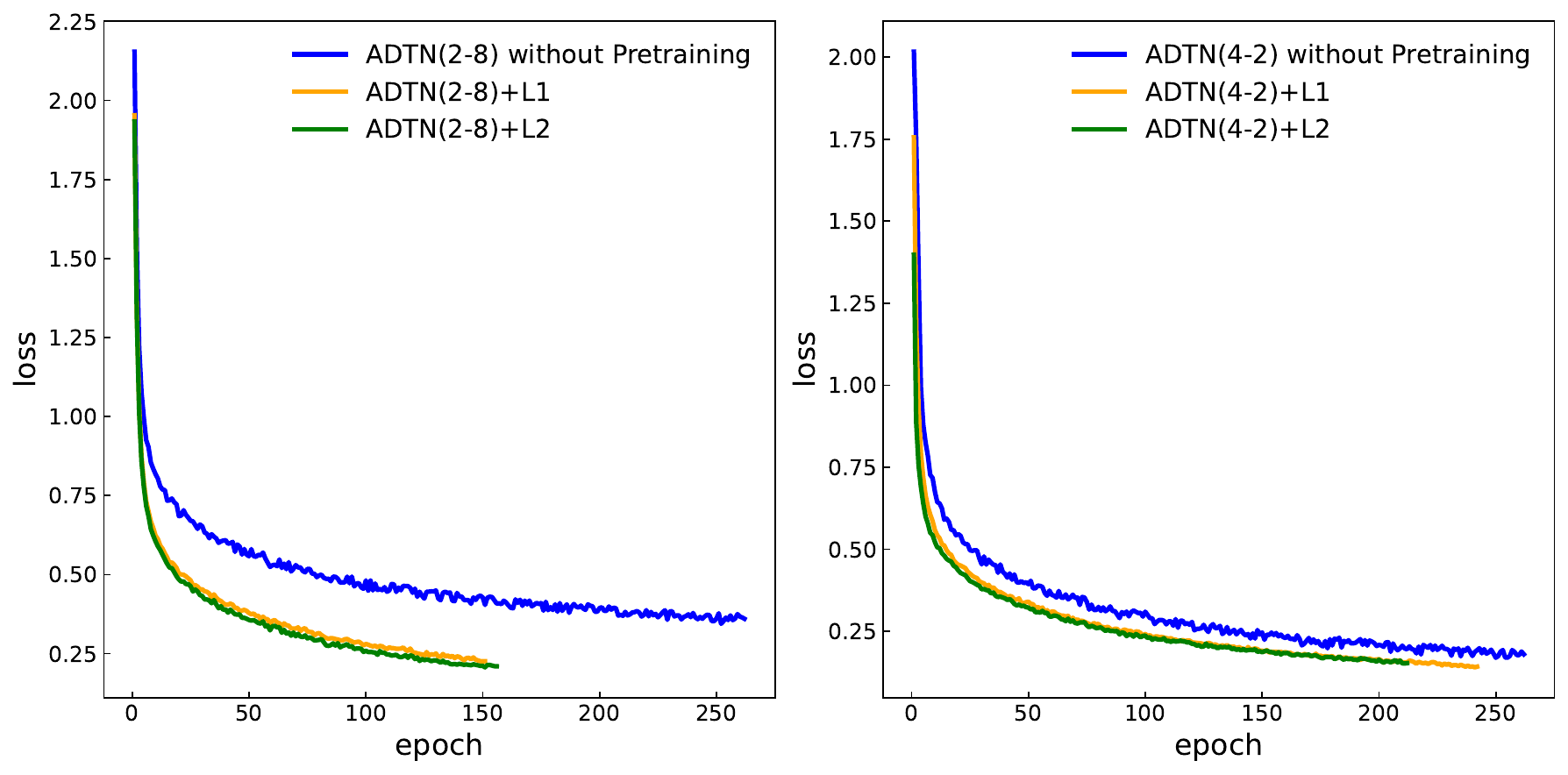}
	\label{fig-loss}
\end{figure*}

\R{Fig.~\ref{fig-loss} compares the losses versus the training epochs during the fine-tuning stage, with or without pre-training. The results using the L1 and L2 (Euclidean) distances for pre-training are shown. In this experiment, we compress the last three layers of AlexNet simultaneously, excluding the output layer.} Faster convergence and lower losses are achieved with pre-training. Meanwhile, the L2 distance {results in a} slightly lower loss than the L1 distance.

\section{Hyperparameters of ADTN's for the compressing VGG-16}

In the whole work, we fix the dimensions of each tensor in the ADTN's to be \((2 \times 2 \times 2 \times 2)\), and take the brick-wall TN structure. The other key hyperparameters of the ADTN compression scheme are the number of ADTN's, $M$, and the value of $Q$ for each ADTN (i.e.,  the order of tensor obtained by contracting the ADTN). Table \ref{tab:smvgg} shows the values of $Q$ for the compression of the NN layers in VGG-16. We fix $M=2$. Each ADTN is configured with a brick-wall architecture, as shown in Fig. \ref{fig-aTN}. For an NN layer, the number of parameters compressed by the two ADTN's is given by \(2^{Q_1} + 2^{Q_2}\), where the values of $Q$ (denoted as $Q_1$ and $Q_2$ for the two ADTN's, respectively) are taken as large and as equal as possible. Correspondingly, the number of parameters in the two ADTN's is
\[
16 M (Q_1 + Q_2 - 2).
\]

\renewcommand{\arraystretch}{1.7}
\begin{table}[htp]
    \caption{\R{Sample of compressed VGG-16 with ADTN(N-M), where $N=2$, $M=4$.}}
    \label{tab:smvgg}
    \centering
    \Large 
    \resizebox{\linewidth}{!}{
\begin{tabular}{cccccc}
\hline
\textbf{Layer name} & \textbf{Weight shape}            & \textbf{Params} & \textbf{$Q_1$} & \textbf{$Q_2$} & \textbf{ADTN Params} \\ \hline
Conv1               & 96$\times$3$\times$3$\times$3    & -               & -              & -              & -                    \\ \hline
Conv2               & 96$\times$96$\times$3$\times$3   & -               & -              & -              & -                    \\ \hline
Conv3               & 128$\times$96$\times$3$\times$3  & -               & -              & -              & -                    \\ \hline
Conv4               & 128$\times$128$\times$3$\times$3 & $2^{17}+2^{14}$ & $17$           & $14$           & $1856$               \\ \hline
Conv5               & 256$\times$128$\times$3$\times$3 & $2^{18}+2^{15}$ & $18$           & $15$           & $1984$               \\ \hline
Conv6               & 256$\times$256$\times$3$\times$3 & $2^{19}+2^{16}$ & $19$           & $16$           & $2112$               \\ \hline
Conv7               & 256$\times$256$\times$3$\times$3 & $2^{19}+2^{16}$ & $19$           & $16$           & $2112$               \\ \hline
Conv8               & 512$\times$256$\times$3$\times$3 & $2^{20}+2^{17}$ & $20$           & $17$           & 2240                 \\ \hline
Conv9               & 512$\times$512$\times$3$\times$3 & $2^{21}+2^{18}$ & $21$           & $18$           & 2368                 \\ \hline
Conv10              & 512$\times$512$\times$3$\times$3 & $2^{21}+2^{18}$ & $21$           & $18$           & 2368                 \\ \hline
Conv11              & 512$\times$512$\times$3$\times$3 & $2^{21}+2^{18}$ & $21$           & $18$           & 2368                 \\ \hline
Conv12              & 512$\times$512$\times$3$\times$3 & $2^{21}+2^{18}$ & $21$           & $18$           & 2368                 \\ \hline
Conv13              & 512$\times$512$\times$3$\times$3 & $2^{21}+2^{18}$ & $21$           & $18$           & 2368                 \\ \hline
FC1                 & 512$\times$4096                  & $2^{20}+2^{20}$ & $20$           & $20$           & 2304                 \\ \hline
FC2                 & 4096$\times$4096                 & $2^{23}+2^{23}$ & $23$           & $23$           & 2688                 \\ \hline
FC3                 & 4096$\times$class                & -               & -              & -              & -                    \\ \hline
\end{tabular}%
}
\end{table}

\section{Basics of tensor network and the contractions}

\label{app-A}

\begin{figure}[htp]
	\centering
	\includegraphics[angle=0,width=1\linewidth]{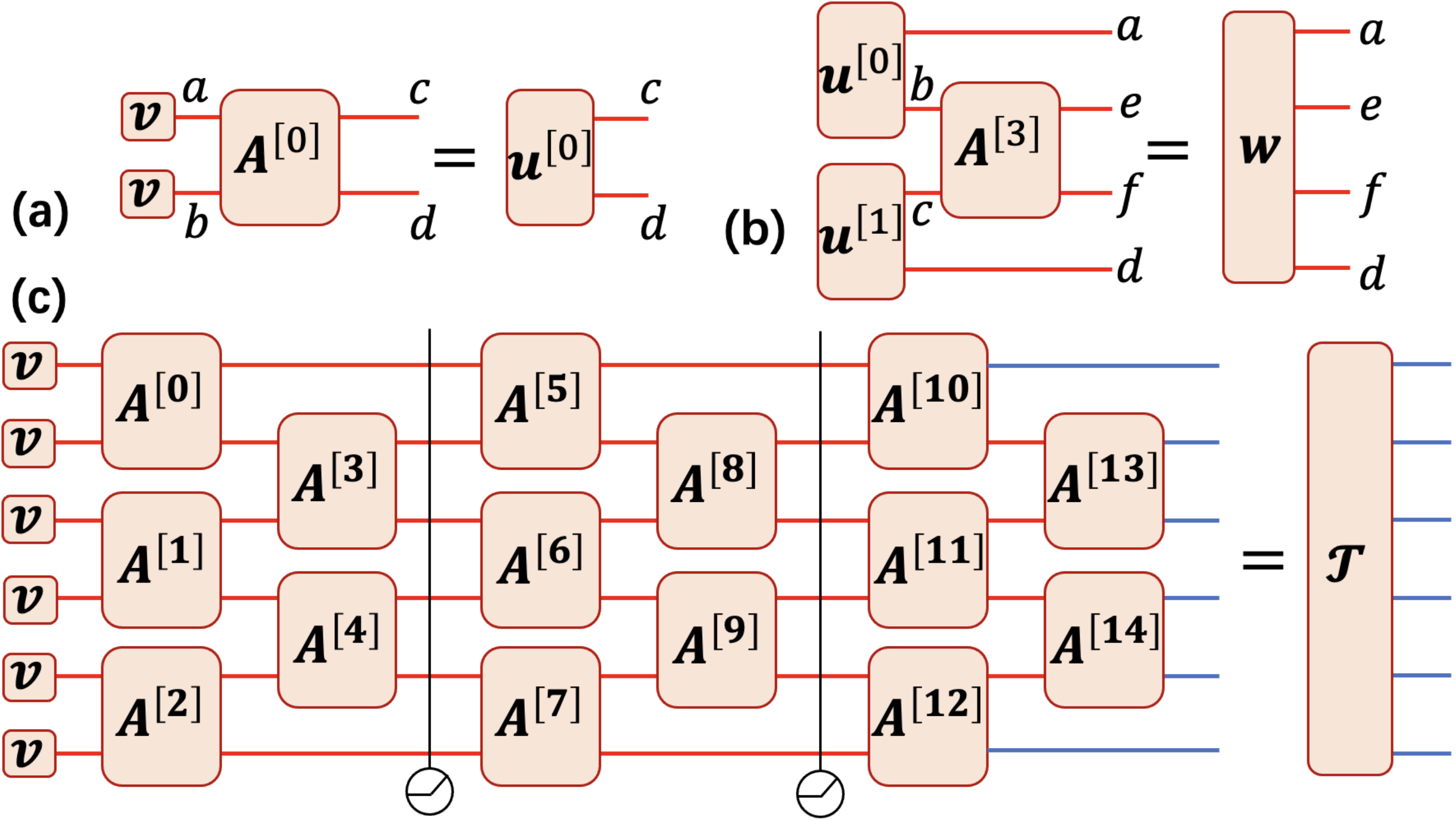}
	\caption{(Color online) In (a) and (b), we show the diagrammatic representations of two local contractions in the ADTN. The contraction of the whole ADTN results in a higher-order tensor $\boldsymbol{\mathcal{T}}$ that stores the variational parameters of NN. The lowercase letters around the bonds indicate the indexes of tensors.}
	\label{fig-aTN}
\end{figure}

To explain how the higher-order tensor $\boldsymbol{\mathcal{T}}$ by contracting the ADTN, we show in Fig.~\ref{fig-aTN} (a) and (b) the diagrammatic representations of two local contractions as an example. As shown in (a), the contraction of one tensor $\boldsymbol{A}^{[0]}$ in ADTN and two vectors $\boldsymbol{v}$ reads
\begin{align}
	u^{[0]}_{cd} = \sum_{ab} v_{a} v_{b} A^{[0]}_{abcd}. 
	\label{eq-contract1}
\end{align}
Another example is the contraction shown in (b), which reads
\begin{align}
	w_{aefd} = \sum_{bc} u^{[0]}_{ab} u^{[1]}_{cd} A^{[3]}_{bcef},
	\label{eq-contract2}
\end{align}
where $\boldsymbol{u}^{[1]}$ can be the result of the contraction of $\boldsymbol{A}^{[1]}$ and two $\boldsymbol{v}$'s. Generally speaking, the contraction formulas and their diagrammatic representations are in one-to-one correspondence. After contracting all shared bonds (indexes) by following the same rules as Eqs.~(\ref{eq-contract1}) and (\ref{eq-contract2}), the ADTN results in a higher-order tensor $\boldsymbol{\mathcal{T}}$ that represents the variational parameters of NN, whose indexes are illustrated by the blue bonds in Fig.~\ref{fig-aTN} (c).

With the presence of non-linear activations, the contractions should be done layer by layer, from left to right. After contracting one tensor layer, each element of the obtained tensor will be put into the activation function. In the simulations, we take the ReLU activation, where an element $x$ is mapped as
\begin{align}
	\sigma(x)=\max (0, x).
\end{align}

\section{Details of FC-2, LeNet-5, AlexNet, AFNet and VGG-16}

Below, we provide the structure and hyper-parameters of the NN's (FC2 in Table \ref{tab:fc2}, LeNet-5(MNIST) in Table \ref{tab:lenet(MNIST)}, LeNet-5 in Table \ref{tab:lenet}, AlexNet in Table \ref{tab:AlexNet}, VGG-16 in Table \ref{tab:vgg16} and reduced version VGG-16 in Table \ref{tab:reduced version vgg16}), for the convenience of reproducing our results. 

\begin{table}[htp]
	\begin{center}
		\caption{Details of FC2.}
		\label{tab:fc2}
		\resizebox{0.8\linewidth}{!}{
			\begin{tabular}{ccccc}
				\textbf{Layer name} & \textbf{Input size} & \textbf{Paras} & \textbf{Output size} \\
				\hline
				FC1 & 784 &$784\times256$ & 256\\
				\hline
				\multicolumn{4}{c}{ReLU}\\
				\hline
				FC2 & 256 &$256\times10$ & 10  \\
				\hline
		\end{tabular}}
	\end{center}
\end{table}

\begin{table}[htp]
	\begin{center}
		\caption{Details of LeNet-5(MNIST).}
		\label{tab:lenet(MNIST)}
		\renewcommand\arraystretch{1}
		\resizebox{0.9\linewidth}{!}{
			\begin{tabular}{ccccc}
				\textbf{Layer name} & \textbf{Input size} & \textbf{Paras} & \textbf{Output size}\\
				\hline
				Conv1 & $28\times28\times1$ & [1;6;5;2] & $28\times28\times6$ \\
				\hline
				\multicolumn{5}{c}{ReLU}\\
				\hline
				MaxPool2 & $28\times28\times6$ &  & $14\times14\times6$ &  \\
				\hline
				Conv2 & $14\times14\times6$ & [6;16;5;0] & $10\times10\times16$ \\
				\hline
				\multicolumn{5}{c}{ReLU}\\
				\hline
				MaxPool2 & $10\times10\times16$ &  & $5\times5\times16$ &  \\
				\hline
				Conv3 & $5\times5\times16$ & [16;120;5;0] & $1\times1\times120$ \\
				\hline
				FC1 & 120 & $120\times84$ & 84 \\
				\hline
				\multicolumn{5}{c}{ReLU}\\
				\hline
				FC2 & 84 & $84\times10$ & 10 \\
				\hline
				
		\end{tabular}}
	\end{center}
\end{table}

\begin{table}[htp]
	\begin{center}
		\caption{Details of LeNet-5.}
		\label{tab:lenet}
		\renewcommand\arraystretch{1}
		\resizebox{\linewidth}{!}{
			\begin{tabular}{ccccc}
				\textbf{Layer name} & \textbf{Input size} & \textbf{Paras} & \textbf{Output size}  \\
				\hline
				Conv1 & $32\times32\times3$ & [3;16;5;1;0] & $28\times28\times16$ \\
				\hline
				\multicolumn{5}{c}{ReLU}\\
				\hline
				MaxPool2 & $28\times28\times16$ &  & $14\times14\times16$ &  \\
				\hline
				Conv2 & $14\times14\times16$ & [16;32;5;1;0] & $10\times10\times32$ \\
				\hline
				\multicolumn{5}{c}{ReLU}\\
				\hline
				AdaptiveAvgPool2 & $10\times10\times32$ &  & $4\times4\times32$ &  \\
				\hline
				FC1 & 512 & $512\times256$ & 256 \\
				\hline
				\multicolumn{5}{c}{ReLU}\\
				\hline
				FC2 & 256 & $256\times128$ & 128 \\
				\hline
				\multicolumn{5}{c}{ReLU}\\
				\hline
				FC3 & 128 & $128\times${class} & {class} \\
				\hline
				
		\end{tabular}}
	\end{center}
\end{table}

\begin{table}[htp]
	\begin{center}
		\caption{Details of AlexNet.}
		\label{tab:AlexNet}
		\renewcommand\arraystretch{1}
		\resizebox{\linewidth}{!}{
			\begin{tabular}{ccccc}
				\textbf{Layer name} & \textbf{Input size} & \textbf{Paras} & \textbf{Output size} \\
				\hline
				Conv1 & $32\times32\times3$ & [3;64;3;2;1] & $16\times16\times64$ \\
				\hline
				\multicolumn{5}{c}{BatchNorm2d}\\
				\hline
				\multicolumn{5}{c}{ReLU}\\
				\hline
				MaxPool2 & $16\times16\times64$ &  [3;2] & $7\times7\times64$ &  \\
				\hline
				SpConv2 & $7\times7\times64$ & [64;128;2;1;0] & $7\times7\times128$ \\
				\hline
				\multicolumn{5}{c}{BatchNorm2d}\\
				\hline
				\multicolumn{5}{c}{ReLU}\\
				\hline
				MaxPool2 & $7\times7\times128$ &  [3;2] & $3\times3\times128$ &  \\
				\hline
				SpConv3 & $3\times3\times128$ & [128;256;2;1;0] & $3\times3\times256$ \\
				\hline
				\multicolumn{5}{c}{BatchNorm2d}\\
				\hline
				\multicolumn{5}{c}{ReLU}\\
				\hline
				SpConv4 & $3\times3\times256$ & [256;512;2;1;0] & $3\times3\times512$\\
				\hline
				\multicolumn{5}{c}{BatchNorm2d}\\
				\hline
				\multicolumn{5}{c}{ReLU}\\
				\hline
				SpConv5 & $3\times3\times512$ & [512;512;2;1;0] & $3\times3\times512$ \\
				\hline
				\multicolumn{5}{c}{BatchNorm2d}\\
				\hline
				\multicolumn{5}{c}{ReLU}\\
				\hline
				MaxPool2 & $3\times3\times512$ &  [3;2] & $1\times1\times512$ &  \\
				\hline
				\multicolumn{5}{c}{Dropout(0.5)}\\
				\hline
				FC1 & 512 & $512\times2048$ & 2048 \\
				\hline
				\multicolumn{5}{c}{ReLU}\\
				\hline
				\multicolumn{5}{c}{Dropout(0.5)}\\
				\hline
				FC2 & 2048 & $2048\times2048$ & 2048 \\
				\hline
				\multicolumn{5}{c}{ReLU}\\
				\hline
				FC3 & 2048 & $2048\times${class} & {class} \\
				\hline
		\end{tabular}}
	\end{center}
\end{table}

\begin{table}
	\begin{center}
		\caption{Details of ZFNet.}
		\label{tab:ZF Net}
		\renewcommand\arraystretch{1}
		\resizebox{\linewidth}{!}{
			\begin{tabular}{ccccc}
				\textbf{Layer name} & \textbf{Input size} & \textbf{Paras} & \textbf{Output size}  \\
				\hline
				Conv1 & $32\times32\times3$ & [3;64;3;2;1] & $16\times16\times64$ \\
				\hline
				\multicolumn{5}{c}{ReLU}\\
				\hline
				MaxPool2 & $16\times16\times64$ &   & $8\times8\times64$ &  \\
				\hline
				SpConv2 & $8\times8\times64$ & [64;128;2;1;0] & $8\times8\times128$ \\
				\hline
				\multicolumn{5}{c}{ReLU}\\
				\hline
				MaxPool2 & $8\times8\times128$ &  8 & $4\times4\times128$ &  \\
				\hline
				SpConv3 & $4\times4\times128$ & [128;128;2;1;0] & $4\times4\times128$ \\
				\hline
				\multicolumn{5}{c}{ReLU}\\
				\hline
				SpConv4 & $4\times4\times128$ & [128;256;2;1;0] & $4\times4\times256$ \\
				\hline
				\multicolumn{5}{c}{ReLU}\\
				\hline
				SpConv5 & $4\times4\times256$ & [256;256;2;1;0] & $4\times4\times256$ \\
				\hline
				\multicolumn{5}{c}{ReLU}\\
				\hline
				MaxPool2 & $4\times4\times256$ &   & $2\times2\times256$ &  \\
				\hline
				\multicolumn{5}{c}{Dropout(0.5)}\\
				\hline
				FC1 & 1024 & $1024\times2048$ & 2048 \\
				\hline
				\multicolumn{5}{c}{ReLU}\\
				\hline
				\multicolumn{5}{c}{Dropout(0.5)}\\
				\hline
				FC2 & 2048 & $2048\times2048$ & 2048 \\
				\hline
				\multicolumn{5}{c}{ReLU}\\
				\hline
				\multicolumn{5}{c}{Dropout(0.5)}\\
				\hline
				FC3 & 2048 & $2048\times${class} & {class} \\
				\hline
		\end{tabular}}
	\end{center}
\end{table}

\begin{table}[htp]
	\begin{center}
		\caption{Details of VGG-16.}
		\label{tab:vgg16}
		\renewcommand\arraystretch{1}
		\resizebox{\linewidth}{!}{
			\begin{tabular}{ccccc}
				\textbf{Layer name} & \textbf{Input size} & \textbf{Paras} & \textbf{Output size} \\
				\hline
				Conv1 & $32\times32\times3$ & [3;96;3;1;1] & $32\times32\times96$ \\
				\hline
				\multicolumn{5}{c}{BatchNorm2d}\\
				\hline
				\multicolumn{5}{c}{ReLU}\\
				\hline
				Conv2 & $32\times32\times96$ & [96;96;3;1;1] & $32\times32\times96$ \\
				\hline
				\multicolumn{5}{c}{BatchNorm2d}\\
				\hline
				\multicolumn{5}{c}{ReLU}\\
				\hline
				MaxPool2 & $32\times32\times96$ &  [2;2] & $16\times16\times96$ &  \\
				\hline
				Conv3 & $16\times16\times96$ & [96;128;3;1;1] & $16\times16\times128$ \\
				\hline
				\multicolumn{5}{c}{BatchNorm2d}\\
				\hline
				\multicolumn{5}{c}{ReLU}\\
				\hline
				
				Conv4 & $16\times16\times128$ & [128;128;3;1;1] & $16\times16\times128$ \\
				\hline
				\multicolumn{5}{c}{BatchNorm2d}\\
				\hline
				\multicolumn{5}{c}{ReLU}\\
				\hline
				MaxPool2 & $16\times16\times128$ &  [2;2] & $8\times8\times128$ &  \\
				\hline
				Conv5 & $8\times8\times128$ & [128;256;3;1;1] & $8\times8\times256$ \\
				\hline
				\multicolumn{5}{c}{BatchNorm2d}\\
				\hline
				\multicolumn{5}{c}{ReLU}\\
				\hline
				
				Conv6 & $8\times8\times256$ & [256;256;3;1;1] & $8\times8\times256$ \\
				\hline
				\multicolumn{5}{c}{BatchNorm2d}\\
				\hline
				\multicolumn{5}{c}{ReLU}\\
				\hline
				Conv7 & $8\times8\times256$ & [256;256;3;1;1] & $8\times8\times256$ \\
				\hline
				\multicolumn{5}{c}{BatchNorm2d}\\
				\hline
				\multicolumn{5}{c}{ReLU}\\
				\hline
				MaxPool2 & $8\times8\times256$ &  [2;2] & $4\times4\times256$ &  \\
				\hline
				Conv8 & $4\times4\times256$ & [256;512;3;1;1] & $4\times4\times512$ \\
				\hline
				\multicolumn{5}{c}{BatchNorm2d}\\
				\hline
				\multicolumn{5}{c}{ReLU}\\
				\hline
				Conv9 & $4\times4\times512$ & [512;512;3;1;1] & $4\times4\times512$ \\
				\hline
				\multicolumn{5}{c}{BatchNorm2d}\\
				\hline
				\multicolumn{5}{c}{ReLU}\\
				\hline
				Conv10 & $4\times4\times512$ & [512;512;3;1;1] & $4\times4\times512$ \\
				\hline
				\multicolumn{5}{c}{BatchNorm2d}\\
				\hline
				\multicolumn{5}{c}{ReLU}\\
				\hline
				MaxPool2 & $4\times4\times512$ &  [2;2] & $2\times2\times512$ &  \\
				\hline
				Conv11 & $2\times2\times512$ & [512;512;3;1;1] & $2\times2\times512$ \\
				\hline
				\multicolumn{5}{c}{BatchNorm2d}\\
				\hline
				\multicolumn{5}{c}{ReLU}\\
				\hline
				Conv12 & $2\times2\times512$ & [512;512;3;1;1] & $2\times2\times512$ \\
				\hline
				\multicolumn{5}{c}{BatchNorm2d}\\
				\hline
				\multicolumn{5}{c}{ReLU}\\
				\hline
				Conv13 & $2\times2\times512$ & [512;512;3;1;1] & $2\times2\times512$ \\
				\hline
				\multicolumn{5}{c}{BatchNorm2d}\\
				\hline
				\multicolumn{5}{c}{ReLU}\\
				\hline
				MaxPool2 & $2\times2\times512$ &  [2;2] & $1\times1\times512$ &  \\
				\hline
				
				FC1 & 512 & $512\times4096$ & 4096 \\
				\hline
				\multicolumn{5}{c}{ReLU}\\
				\hline
				\multicolumn{5}{c}{Dropout(0.4)}\\
				\hline
				FC2 & 4096 & $4096\times4096$ & 4096 \\
				\hline
				\multicolumn{5}{c}{ReLU}\\
				\hline
				\multicolumn{5}{c}{Dropout(0.4)}\\
				\hline
				FC3 & 4096 & $4096\times${class} & {class} \\
				\hline
		\end{tabular}}
	\end{center}
\end{table}

\begin{table}[htp]
	\begin{center}
		\caption{Details of VGG-16 (reduced version).}
		\label{tab:reduced version vgg16}
		\renewcommand\arraystretch{1}
		\resizebox{\linewidth}{!}{
			
			\begin{tabular}{ccccc}
				\textbf{Layer name} & \textbf{Input size} & \textbf{Paras} & \textbf{Output size}  \\
				\hline
				Conv1 & $32\times32\times3$ & [3;64;2;1] & $32\times32\times64$ \\
				\hline
				Conv2 & $32\times32\times64$ & [64;64;2;0] & $32\times32\times64$ \\
				\hline
				\multicolumn{5}{c}{ReLU}\\
				\hline
				MaxPool2 & $32\times32\times64$ &  & $16\times16\times64$ &  \\
				\hline
				
				Conv3 & $16\times16\times64$ & [64;128;2;1] & $16\times16\times128$ \\
				\hline
				Conv4 & $16\times16\times128$ & [128;128;2;0] & $16\times16\times128$ \\
				\hline
				\multicolumn{5}{c}{ReLU}\\
				\hline
				MaxPool2 & $16\times16\times128$ &  & $8\times8\times128$ &  \\
				\hline
				
				Conv5 & $8\times8\times128$ & [128;256;2;1] & $8\times8\times256$ \\
				\hline
				Conv6 & $8\times8\times256$ & [256;256;2;1] & $8\times8\times256$ \\
				\hline
				\multicolumn{5}{c}{ReLU}\\
				\hline
				Conv7 & $8\times8\times256$ & [256;256;2;0] & $8\times8\times256$ \\
				\hline
				\multicolumn{5}{c}{ReLU}\\
				\hline
				MaxPool2 & $8\times8\times256$ &  & $4\times4\times256$ &  \\
				\hline
				
				Conv8 & $4\times4\times256$ & [256;512;2;1] & $4\times4\times512$\\
				\hline
				Conv9 & $4\times4\times256$ & [512;512;2;1] & $4\times4\times512$ \\
				\hline
				\multicolumn{5}{c}{ReLU}\\
				\hline
				Conv10 & $4\times4\times256$ & [512;512;2;0] & $4\times4\times512$ \\
				\hline
				\multicolumn{5}{c}{ReLU}\\
				\hline
				MaxPool2 & $4\times4\times512$ &  & $2\times2\times512$ &  \\
				\hline
				
				Conv11 & $2\times2\times512$ & [512;1024;2;1] & $2\times2\times1024$ \\
				\hline
				Conv12 & $2\times2\times1024$ & [1024;1024;2;1] & $2\times2\times1024$ \\
				\hline
				\multicolumn{5}{c}{ReLU}\\
				\hline
				Conv13 & $2\times2\times1024$ & [1024;1024;2;0] & $2\times2\times1024$ \\
				\hline
				\multicolumn{5}{c}{ReLU}\\
				\hline
				
				MaxPool2 & $2\times2\times1024$ &  & $1\times1\times1024$ &  \\
				\hline
				
				\multicolumn{5}{c}{Dropout(0.8)}\\
				\hline
				FC1 & 1024 & $1024\times512$ & 512 \\
				\hline
				\multicolumn{5}{c}{ReLU}\\
				\hline
				\multicolumn{5}{c}{Dropout(0.8)}\\
				\hline
				
				FC2 & 512 & $512\times256$ & 256 \\
				\hline
				\multicolumn{5}{c}{ReLU}\\
				\hline
				\multicolumn{5}{c}{Dropout(0.8)}\\
				\hline
				FC3 & 256 & $256\times10$ & {10} \\
				\hline
		\end{tabular}}
	\end{center}
\end{table}

\end{document}